\documentclass[sigconf]{acmart}

\AtBeginDocument{%
  \providecommand\BibTeX{{%
    \normalfont B\kern-0.5em{\scshape i\kern-0.25em b}\kern-0.8em\TeX}}}

\acmYear{2024}\copyrightyear{2024}
\setcopyright{acmlicensed}
\acmConference[ACM FAccT '24]{ACM Conference on Fairness, Accountability, and Transparency}{June 3--6, 2024}{Rio de Janeiro, Brazil}
\acmBooktitle{ACM Conference on Fairness, Accountability, and Transparency (ACM FAccT '24), June 3--6, 2024, Rio de Janeiro, Brazil}
\acmDOI{10.1145/3630106.3658990           }
\acmISBN{979-8-4007-0450-5/           24/06}

\usepackage{cleveref}
\usepackage{graphicx}
\usepackage{tabularx}
\usepackage{array}
\usepackage{multicol}
\usepackage{multirow}

\begin{document}

\setlength{\fboxsep}{5pt}

\title{A Critical Survey on Fairness Benefits of Explainable AI}

\author{Luca Deck}
\email{luca.deck@uni-bayreuth.de}
\orcid{0000-0003-3773-2769}
\affiliation{
    \institution{University of Bayreuth \& Fraunhofer FIT}
    \city{Bayreuth}
    \country{Germany}
}

\author{Jakob Schoeffer}
\email{schoeffer@utexas.edu}
\orcid{0000-0003-3705-7126}
\affiliation{
    \institution{University of Texas at Austin}
    \city{Austin}
    \country{USA}
}

\author{Maria De-Arteaga}
\email{dearteaga@utexas.edu}
\orcid{0000-0003-2297-3308}
\affiliation{
    \institution{University of Texas at Austin}
    \city{Austin}
    \country{USA}
}

\author{Niklas Kühl}
\email{kuehl@uni-bayreuth.de}
\orcid{0000-0001-6750-0876}
\affiliation{
    \institution{University of Bayreuth \& Fraunhofer FIT}
    \city{Bayreuth}
    \country{Germany}
}

\begin{abstract}
In this critical survey, we analyze typical claims on the relationship between explainable AI (XAI) and fairness to disentangle the multidimensional relationship between these two concepts. Based on a systematic literature review and a subsequent qualitative content analysis, we identify seven archetypal claims from 175 scientific articles on the alleged fairness benefits of XAI. We present crucial caveats with respect to these claims and provide an entry point for future discussions around the potentials and limitations of XAI for specific fairness desiderata. Importantly, we notice that claims are often $(i)$ vague and simplistic, $(ii)$ lacking normative grounding, or $(iii)$ poorly aligned with the actual capabilities of XAI. We suggest to conceive XAI not as an ethical panacea but as one of many tools to approach the multidimensional, sociotechnical challenge of algorithmic fairness. Moreover, when making a claim about XAI and fairness, we emphasize the need to be more specific about \textit{what} kind of XAI method is used, \textit{which} fairness desideratum it refers to, \textit{how} exactly it enables fairness, and \textit{who} is the stakeholder that benefits from XAI.
\end{abstract}



\maketitle

\section{Introduction}

The integration of artificial intelligence (AI) into decision-making processes has raised concerns about reinforcing societal inequalities \citep{Angwin.2022,Ntoutsi.2020}. 
Moreover, much progress in AI comes at the cost of increased complexity and opacity, which may impede human understanding \citep{Burrell.2016}.
Explainable AI (XAI) is commonly conceived as a remedy to both of these challenges \citep{BarredoArrieta.2020}.
However, the implicit link between XAI and fairness has been challenged due to inconclusive evidence and a lack of consistent terminology \citep{Langer.2021b, Balkir.2022}. 

Our critical survey explores the complex relationship between XAI and algorithmic fairness by reviewing 175 recent articles and identifying seven archetypal claims on the alleged fairness benefits of XAI.
Organizing the scattered debate into meaningful sub-debates, we discuss caveats and provide an entry point for future discussions on the suitability and limitations of XAI for fairness.
We find that literature from various domains is highly optimistic about the usefulness of XAI for several fairness dimensions and stakeholders.
However, many claims in the literature remain vague about how exactly XAI will contribute to fairness.
They disregard technical limitations, conflicts of interest between stakeholders, and normative grounding.
Highlighting central caveats as well as nascent approaches to address these caveats, we contribute to a more nuanced understanding of the interplay between XAI and fairness in AI-informed decision-making.

This article is structured as follows:
we start by establishing key concepts that we use to structure the debate and interpret claims.
Next, we describe how we identified and organized claims from 175 articles into 7 archetypal categories.
Afterwards, we introduce each archetypal claim at face value by verbalizing the underlying intuition from the literature.
Based on that, we organize a meaningful, structured debate and take a critical perspective on these claims.
Finally, we synthesize patterns of critique and caveats to embed the alleged fairness benefits of XAI in a bigger picture.
Through this thorough and systematic study, we hope to bring clarity to the entangled relationship between XAI and fairness and inform future efforts aimed at leveraging XAI to tackle algorithmic unfairness.

\section{Background}
\label{sec:back}

\paragraph{Related work}
XAI and fairness have both separately produced a large body of research~\citep{Speith.2022,caton2020fairness}. Recent work has noted that while claims regarding the intersection are frequent, there is a lack of specificity regarding how they relate \citep{Meng.2022, Balkir.2022}.
\citet{Langer.2021b} extract different desiderata of XAI (with one of them being fairness) from over 100 peer-reviewed publications and point out that only a subset of articles substantiates their claims with empirical evidence.
\citet{Balkir.2022} survey challenges in the application of XAI for fairer language models and call for more precise conceptualization on how exactly XAI relates to fairness.
\citet{Zhou.2020b} provide an overview of XAI methods for fairness objectives and highlight the consideration of contextual factors as well as the need for interdisciplinary research.
Our work adds to this stream of literature by adopting a systematic and critical approach similar to \citet{Blodgett.2020}, who critically analyze the use of the term ``bias'' in the context of natural language processing.

\paragraph{Dimensions of XAI and fairness}
Both XAI and fairness are multifaceted concepts that can be conceptualized along various dimensions~\citep{Speith.2022,caton2020fairness}.
In this work, we understand XAI as any mechanism that ``produces details or reasons to make [the] functioning [of an AI system] clear or easy to understand''~\citep{BarredoArrieta.2020}.
We adopt the term \textit{desideratum} from \citet{Langer.2021b} to differentiate several roles and objectives of XAI.
We explicitly examine desiderata in the context of fairness and distinguish between formalized notions of fairness (\textit{formal fairness}) and human perceptions of fairness (\textit{perceived fairness}), similar to \citet{Starke.2022}.
Formal fairness criteria are captured in mathematical and statistical frameworks \citep{Barocas.2019, Castelnovo.2022}, which may or may not align with human fairness perceptions \citep{Dodge.2019, Nyarko.2021,Srivastava.2019}.
Formal fairness notions are often distinguished into group and individual fairness: group fairness criteria typically require a form of parity between demographic groups, for example, along sensitive features like gender or race~\citep{Chouldechova.2017}.
Individual fairness criteria typically demand to treat similar people alike~\citep{Dwork.2012}.
Perceived fairness is a subjective human attitude that is highly context-sensitive \citep{Starke.2022} and related to complex moral deliberations \citep{Binns.2018}.
It requires fundamentally different measurements than formal fairness, for example, based on psychological constructs \citep{Colquitt.2001}.
From \citet{Colquitt.2001}, we also adopt the decomposition into \textit{distributive}, \textit{procedural}, and \textit{informational} dimensions.
Distributive fairness is primarily concerned with decision outcomes, whereas procedural fairness refers to the underlying decision-making process.
Informational fairness accounts for aspects of the communication accompanying a decision.

Regarding the purpose of XAI, we further differentiate \textit{substantial} from \textit{epistemic} goals~\citep{Langer.2021b}.
An epistemic goal refers to the capability of humans to \textit{observe} fairness properties of a model (e.g., XAI providing insights into a model's reliance on sensitive features), whereas a substantial goal actively aims to \textit{alter} fairness properties (e.g., mitigating formally unfair model characteristics).
This distinction is helpful in understanding the multifaceted role of XAI across many application contexts.
For example, an epistemic usage of XAI may be to \textit{inform} about a given fairness desideratum (e.g., group fairness), whereas a substantial usage of XAI aims at directly or indirectly \textit{affecting} (un)fairness properties of an AI system.
We also adopt the taxonomy from \citet{Langer.2021b} on human stakeholders of XAI, which includes developers, deployers, regulators, users, and affected parties (i.e., decision subjects).

\begin{figure*}[t]
    \centering
    \includegraphics[width=0.9\textwidth]{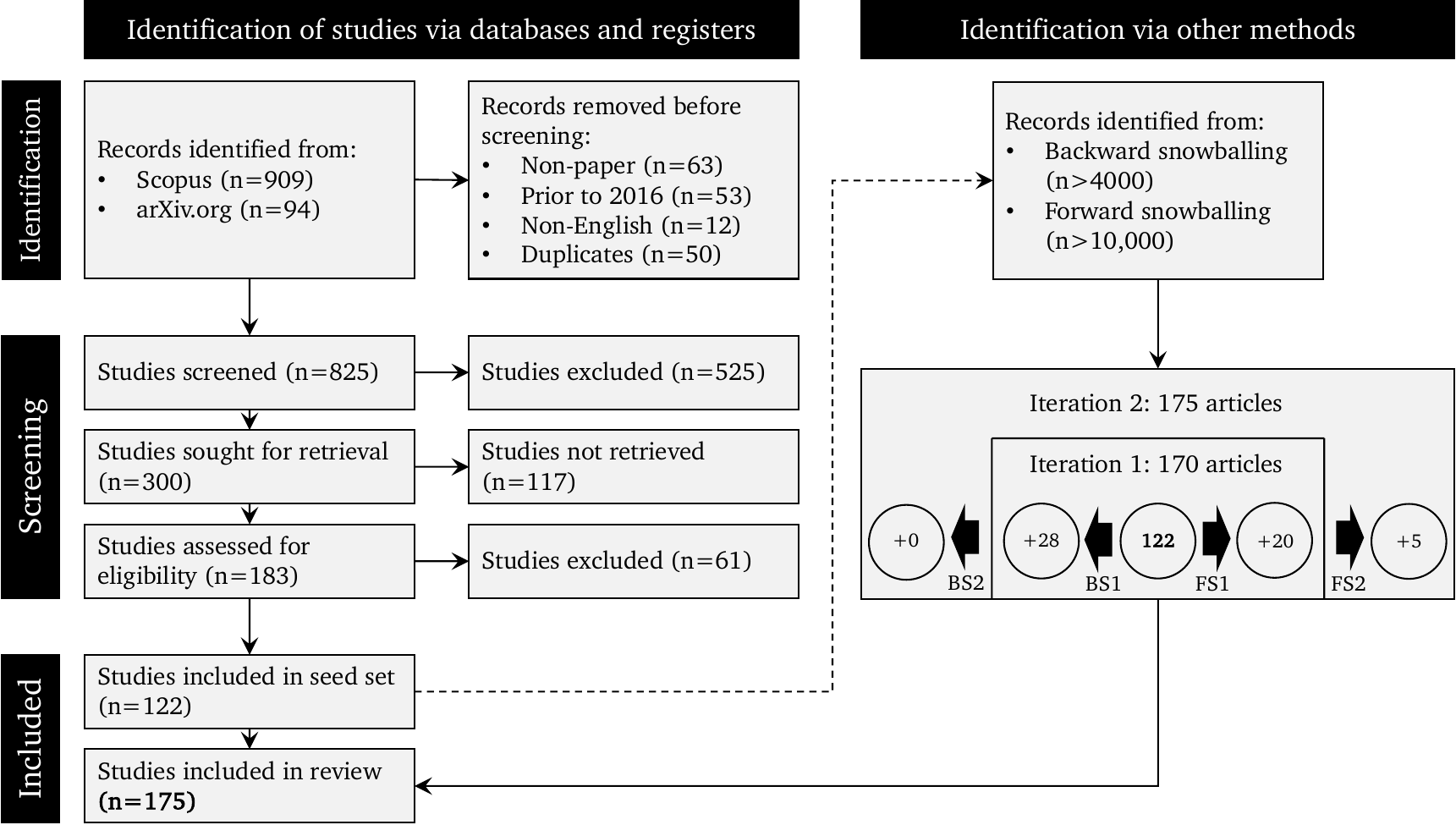}
    \caption{PRISMA flowchart describing the article selection procedure.}
    \label{fig:litrev}
\end{figure*}

\section{Methodology}
Similar to \citet{Blodgett.2020}, we systematically identified and scrutinized claims of recent publications---in our case, about the fairness benefits of XAI.
We first conducted a structured literature review guided by \citet{Kitchenham.2007} to identify entrenched claims on the alleged capabilities of XAI for fairness.
This process yielded 175 articles and is depicted in \Cref{fig:litrev}.
We supplemented our deductive literature review with inductive coding~\citep{Wolfswinkel.2013} at the level of individual claims.
A rigorous qualitative analysis of these claims using a grounded theory approach \citep{ChunTie.2019} yielded seven archetypal claims, summarized in \Cref{fig:claims}.

\subsection{Systematic Literature Review}\label{subsec:systematic}

In order to gain an understanding of the domain, test the effectiveness of keywords, and identify relevant publishers, we initiated an exploratory review by crawling the Google Scholar database. 
Our search string is chosen to reflect various dimensions of both XAI and fairness and to restrict the results to AI contexts. 
For XAI, we relied on the taxonomy of \citet{BarredoArrieta.2020} and also incorporated the related terms \textit{understandability, comprehensibility, interpretability, explainability, and transparency}, as well as their inflections of the same stem.
After screening around 400 individual articles, we finally decided on the following search string---note that the asterisk as a wildcard character allows us in each case to consider words of the same stem, including adjectives and nouns:

\begin{center}
\noindent
\fbox{%
    \parbox{0.8\columnwidth}{%
    \centering
    \big(xai \textbf{OR} explanation \textbf{OR} understandab* \textbf{OR} intelligib* \textbf{OR} comprehensib* \textbf{OR} interpretab* \textbf{OR} explainab* \textbf{OR} transparen*\big)
    \textbf{AND} fair*
    \textbf{AND} \big(ai \textbf{OR} ``artificial intelligence'' \textbf{OR} ``machine learning''\big)
    }
    }
\end{center}

Relying on recent recommendations to combine two popular search strategies, database querying and snowballing \citep{Wohlin.2022}, we followed proven guidelines for systematic literature reviews in the domain of software engineering \citep{Kitchenham.2007, Wohlin.2014}.
Scopus was the natural choice for our database search because it has been recommended as an effective tool to generate seed sets for snowballing \citep{Mourao.2020}, and it includes the most relevant publishers for our task.\footnote{\url{https://www.elsevier.com/products/scopus}}
We allow for
snowballing to include publications outside of Scopus using the lens.org database.\footnote{\url{https://www.lens.org}}
To account for recent, unreviewed publications, we also applied our search string to the arXiv.org database.
Following the documentation guidelines of \citet{Kitchenham.2007} and the PRISMA standard \citep{Page.2021}, we provide a transparent and replicable documentation of the selection process: 
\Cref{fig:litrev} depicts how a total body of 1,003 identified records (as of September 2022) was condensed to a seed set of 122 with explanations on the filter criteria for each step.
Because the goal of this critical survey is to characterize the \emph{contemporary} discourse on XAI and fairness, we focused on articles from 2016 onwards.

At the initial identification level, we only considered scientific articles and excluded records such as courses, keynotes, etc. 
Here, we manually inspected all abstracts and only retained articles examining dimensions of both XAI \textit{and} fairness fitting into the broader scheme of our definitions. 
Consequently, we discarded articles having too broad or deviating notions of the XAI terms (e.g., using the term \textit{explain} in a different context), articles using the term \textit{fair} in different contexts (\textit{fairly, FAIR principles}, etc.), and articles where fairness or XAI are not the object of research.
Proceeding to full-text analyses, we heuristically scanned the entire article for explicit claims on XAI and fairness.
We focused on unique statements as opposed to straightforward summaries or paraphrases of previous work, which eliminated most literature reviews.
Finally, we discarded articles where the direct relationship between XAI and fairness was not considered or remained too vague to infer any claims.
For example, \citet{Shin.2020} examines the influence of explainability and fairness on trustworthiness but does not address the interaction of explainability and fairness.

Starting from the seed set of 122 articles, we performed iterative backward and forward snowballing \citep{Wohlin.2014}. 
Using the citation crawling tool Citationchaser \citep{Haddaway.2022}, we accelerated the snowballing procedure and focused on the most frequently referenced articles.
We stopped whenever the third iteration did not generate any further hits.
Please refer to \Cref{sub:overview} for a comprehensive overview of the gathered articles categorized according to their underlying methodology.

\subsection{Inductive Claim Analysis}\label{subsec:claim}

In a subsequent step, we inductively identified dominant themes by analyzing commonalities and grouping claims into meaningful categories.
We found grounded theory to be an appropriate methodology and followed the research design framework by \citet{ChunTie.2019}, employing MAXQDA for claim extraction, coding, and memoing \citep{Kuckartz.2019}.
To reiterate, we built our grounded theory around the following two questions: 
\begin{itemize}
    \item What does recent literature claim about the relationship between XAI and fairness?
    \item On what kind of evidence are these claims grounded?
\end{itemize}

\begin{figure*}[t]
    \centering
    \includegraphics[width=0.9\linewidth]{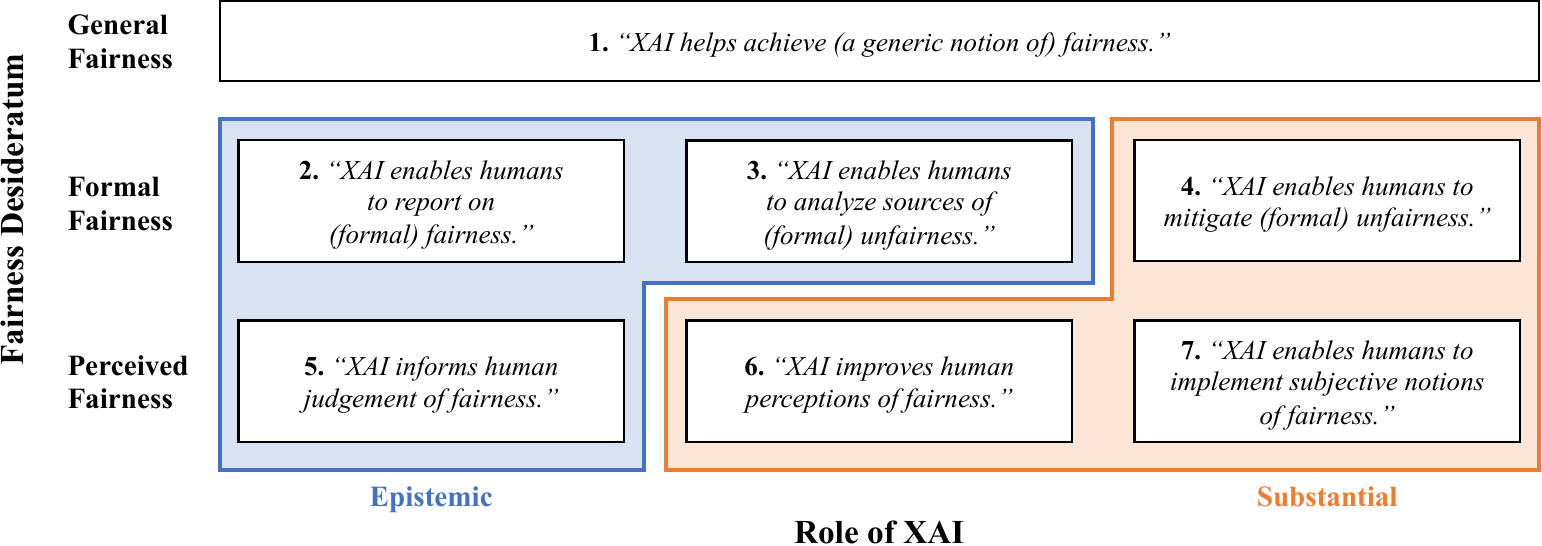}
    \caption{The seven identified archetypal claims on the role of XAI for fairness desiderata.}
    \label{fig:claims}
\end{figure*}

We started by skimming the full texts of the 175 selected articles to comprehend the respective methodology and key results.
In parallel, we scanned for claims with a strong focus on the most relevant and promising article sections. 
For example, the introduction, discussion, and conclusion sections often provided more expressive claims than the method and results sections, which we usually only considered to retrace methodology or reasoning.
Throughout the coding procedure, we used memos to note down important insights, augment the claims with contextual information (such as textual context, meaning of abbreviations, authors’ reasoning, etc.), and document the coders' line of thought.

In the first iteration, we kept the codes as specific as possible to maintain a maximum amount of information.
During coding, we considered not only the verbatim content of the claims but also their context and, if possible, their underlying reasoning.
We used this information to categorize the type of evidence leading to the claim, which we recorded in our coding system.
In the subsequent iterations, we identified higher-level concepts and started grouping the claims into mutually exclusive categories.
To achieve theoretical saturation, we ensured that the identified categories were sufficient and the assignment was plausible and correct by re-checking each claim~\citep{saunders2018saturation}.

\section{Critical Survey}

\Cref{fig:claims} shows the result of our coding process and highlights the different roles XAI may take to address fairness desiderata.
We use the concepts introduced in \Cref{sec:back} to organize the claims and relate them to one another. 
This allows us to differentiate between claims about formal or perceived fairness and claims on what we call ``a generic notion of fairness,'' which do not specify what kind of fairness is pursued.
We also differentiate whether the claims relate to an epistemic or a substantial role of XAI.

In what follows, we introduce each archetypal claim by verbalizing the underlying intuition and providing representative quotes from the literature.
We then organize a structured debate and take a critical perspective.
Please refer to \Cref{tab:claim_examples} for a comprehensive overview of archetypal claims, including exemplary supportive and critical quotes from the literature.
We also provide a complete list of references per claim in \Cref{sub:claim_overview}.

\begin{table*}[t]
    \small
    \caption{Exemplary supportive and critical perspectives on archetypal claims.}\label{tab:claim_examples}
    \begin{tabularx}{\textwidth}{>{\hsize=0.5\hsize}X >{\hsize=0.3\hsize}X >{\hsize=2.2\hsize}X} 
    \toprule
    \textbf{Archetypal claim} & \textbf{Stance} & \textbf{Exemplary quotes} \\
    \midrule
    \multirow{2}{=}{\textit{``XAI helps achieve (a generic notion of) fairness.''}}
    & \textbf{Supportive}
    & \textit{``Understanding the logic and technical innerworkings (i.e. semantic content) of these systems is a precondition for ensuring their safety and fairness''} \citep[p. 40]{Leslie.2019}.
    \vspace{1mm}\\
    & \textbf{Critical}
    & \textit{``[A] perfectly auditable algorithmic decision [...] can nevertheless cause unfair and transformative effects''}~\citep[pp. 14--15]{Mittelstadt.2016}.\\
    \midrule
    \multirow{2}{=}{\textit{``XAI enables humans to report on (formal) fairness.''}}
    & \textbf{Supportive} 
    & \textit{``Given a classifier, a dataset and a set of sensitive features, [the proposed method] first assesses whether the classifier is fair by checking its reliance on sensitive features using `Lime explanations'\thinspace''} \citep[p. 1]{Alves.2021b}.
    \vspace{1mm}\\
    & \textbf{Critical}
    & \textit{``We found that despite their capabilities in simplifying and explaining model behavior, many prominent XAI tools lack features that could be critical in detecting bias''} \citep[p. 1]{Alikhademi.2021}.\\
    \midrule
    \multirow{2}{=}{\textit{``XAI enables humans to analyze sources of (formal) fairness.''}}
    & \textbf{Supportive} 
    & \textit{``To gain insight into how the model discriminates, we construct a transparency report that summarizes the most salient statistical differences in the features between the flipset individuals and their images under the optimal transport mapping [...] Intuitively, the transparency report can serve as an overview of what features the model may be using to discriminate between populations''} \citep[p. 112]{Black.2020}.
    \vspace{1mm}\\
    & \textbf{Critical}
    & \textit{``[LIME] still lacks the skills to detect issues of biased data and detect issues in the selection or processing of the model''} \citep[p. 12]{Alikhademi.2021}.\\
    \midrule
    \multirow{2}{=}{\textit{``XAI enables humans to mitigate (formal) unfairness.''}}
    & \textbf{Supportive} 
    & \textit{``If deemed unfair, [the proposed XAI-based method] then applies feature dropout to obtain a pool of classifiers. These are then combined into an ensemble classifier that was empirically shown to be less dependent on sensitive features''} \citep[p. 3]{Alves.2021b}.
    \vspace{1mm}\\
    & \textbf{Critical} 
    & \textit{``The best one could do with these [XAI] techniques seem to be to develop `colourblind' models which, even if they receive explicit information about protected attributes in their input, ignore this information when making their decisions. Although it is simple and intuitive, we suspect that such an approach has similar issues with the much criticized `fairness through unawareness' approach''} \citep[p. 8]{Balkir.2022}.\\
    \midrule
    \multirow{2}{=}{\textit{``XAI informs human judgement of fairness.''}}
    & \textbf{Supportive} 
    & \textit{``The results [...] suggest that standards clarity and outcome explanation allowed people to judge whether the fairness properties of the algorithm were in line with their fairness concepts''} \citep[p. 18]{Lee.2019b}.
    \vspace{1mm}\\
    & \textbf{Critical} 
    & \textit{``Since the right to explanation as defined in current regulations [...] does not give precise directives on what it means to provide a `valid explanation' [...], there is a legal loophole that can be used by dishonest companies to cover up the possible unfairness of their black-box models by providing misleading explanations''} \citep[p. 1]{Aivodji.2019}.\\
    \midrule
    \multirow{2}{=}{\textit{``XAI improves human perceptions of fairness.''}}
    & \textbf{Supportive} 
    & \textit{``When the system produces a `positive' output, it may be considered fair and every explanation style (except for demographic-based) works, with a slight advantage to our new [...] explanation''} \citep[p. 9]{ShulnerTal.2022b}.
    \vspace{1mm}\\
    & \textbf{Critical} 
    & \textit{``When explanations helped respondents understand biased distributions, perceived fairness decreased''} \citep[p. 7]{Starke.2022}.\\
    \midrule
    \multirow{2}{=}{\textit{``XAI enables humans to implement subjective notions of fairness.''}}
    & \textbf{Supportive} 
    & \textit{``As a user mentioned, `Feature selection is sometimes arbitrary, but it provides the feature-level measures as evidence of fairness-aware decision.' --- this demonstrated how the system can help decision makers to achieve fair decision making through better explainability''} \citep[p. 9]{Ahn.2020}.
    \vspace{1mm}\\
    & \textbf{Critical} 
    & \textit{``We conjecture that miscalibrated fairness perceptions (e.g., due to misleading explanations) may influence reliance on AI in undesirable ways, by making people adopt incorrect or override correct AI recommendations. This lends support to the hypothesis that there is a disconnect between what explanations provide and the fairness benefits they claim''} \citep[p. 2]{Schoeffer.2022b}.\\
    \bottomrule
    \end{tabularx}
\end{table*}

\subsection{Claim 1: \textit{``XAI helps achieve (a generic notion of) fairness.''}} \label{sub:unspecific}
This type of claim treats fairness as a monolithic concept without specifying \textit{how} XAI will lead to \textit{which} kind of fairness for \textit{whom}.
While phrasing and determinism vary by reference, we identify three tendencies.
The first suggests XAI as a \textit{necessary} condition for fairness.
For example, a popular belief is that a decision has to be understandable in order to be fair.
    \begin{quote}
    \small{
    \textit{``First and most evidently, understanding the logic and technical innerworkings (i.e. semantic content) of these systems is a precondition for ensuring their safety and fairness''}~\citep[p. 40]{Leslie.2019}.
    }
    \end{quote}
XAI is other times treated as \textit{sufficient} for achieving fairness.
An exemplary intuition is that revealing the underlying mechanisms of a decision is all it takes to guarantee fairness.
    \begin{quote}
    \small{
    \textit{``From a social standpoint, explainability can be considered as the capacity to reach and guarantee fairness in ML models''}~\citep[p. 9]{BarredoArrieta.2020}.
    }
    \end{quote}
Many others remain \textit{tentative} insofar as they suggest vague capabilities of XAI for fairness but are less assertive.
\begin{quote}
\small{
\textit{``Explainability and interpretability: these two concepts are seen as possible mechanisms to increase algorithmic fairness, transparency and accountability''}~\citep[p. 2]{Cath.2018}.
}
\end{quote} 
    
\paragraph{\textbf{Critique: Futility and danger of vague claims on generic fairness notions}} \label{subsub:futility}
Fairness is a multidimensional concept \citep{Colquitt.2001, Mulligan.2019} with many conflicting notions \citep{Kleinberg.2017, Friedler.2019}.
Hence, it is easy to come up with counterexamples for such strong and simplistic claims.
For example, distributions of classification rates can be used to show that models conform with formal fairness criteria without XAI methods (e.g., \citep{Chouldechova.2017}).
Moreover, a transparent model can be perfectly scrutable and still be deemed unfair by some stakeholders, which precludes the suggested sufficiency \citep{Mittelstadt.2016}.
The central underlying assumption behind these claims appears to be that XAI is valuable to \textit{some} dimensions of fairness.
This (perhaps plausible) intuition is also reflected in all ethical AI principles reviewed by \citet{Floridi.2018}.
However, we argue that suggesting a universal link between XAI and fairness is misleading and threatens a meaningful debate that should account for the multidimensionality of fairness and incorporate the essential needs of relevant stakeholders \citep{Langer.2021b}.
Perpetuating these overly general claims threatens to produce unwarranted trust and reliance on current XAI technologies---and fails to recognize the multitude of interests with regard to fairness. 
By disentangling the multidimensionality of fairness into more fine-grained scopes, our work organizes the debate in a way that fosters more precise claims and nuanced arguments.

\subsection{Claim 2: \textit{``XAI enables humans to report on (formal) fairness.''}} \label{sub:report}

A large share of articles is concerned with using XAI to measure and report on formal (un)fairness, often phrased as ``identifying bias''~\citep{Gilpin.2018} or ``detecting discrimination''~\citep{Jain.2020}.
One central intuition behind these claims is that conventional evaluation of model outcomes (e.g., testing for demographic parity) may fail to consider the underlying mechanisms leading to this outcome~\citep{Lipton.2018}.
XAI is expected to fill this gap by providing insights into these mechanisms, which may then be related to formal fairness criteria \citep{DoshiVelez.2017}.
Since the anti-discriminatory motivation of formal fairness criteria typically relates to sensitive features such as gender or race, XAI is commonly employed to examine how models make use of these sensitive features.
Inspired by the legal notion of disparate treatment \citep{Lipton.2018b} and the idea of ``fairness through unawareness''~\citep{Kusner.2017}, one form of individual fairness deems any explicit use of sensitive features as unfair (e.g., \citep{Alves.2020, Hall.2019, Ignatiev.2020}).

    \begin{quote}
    \small{
    \textit{``An algorithm is fair if the protected features are not explicitly used in the decision-making process''}~\citep[p. 850]{Ignatiev.2020}.
    }
    \end{quote}
    
Moreover, in cases where we have clear guidance on which features are legitimate and fair to use, prior work has argued that feature importance can be employed to validate whether models actually rely on these features; for example, in computer vision tasks (e.g., \citep{Anders.2022}).
Prominent XAI methods that have been employed to gauge the use of sensitive features are feature importance measurements using LIME~\citep{Ribeiro.2016} and SHAP~\citep{Lundberg.2017} (e.g., \citep{Cesaro.2020, Jain.2020, Alves.2021b}), or inherently interpretable models (e.g., \citep{Raff.2018, Tolan.2019, Meng.2022}).

Another common approach to individual fairness focuses on ``contrastive explanations'' \citep{Sokol.2019} for identifying instances where the outcome changes only based on ``flipping'' a sensitive feature.
Articles differ in their exact methodology but often rely on similarity-based measures (e.g., \citep{Black.2020}) or counterfactually constructed data points (e.g., \citep{Dash.2022}) to identify data pairs where this behavior occurs.

Because sensitive features are often correlated to other (seemingly) legitimate features, some works additionally account for correlations and causal relationships of sensitive features (e.g., \citep{Datta.2016, Grabowicz.2022}).

\paragraph{\textbf{Critique: Misconceptions of feature importance for group fairness}} \label{subsub:group}
Logical conclusions between feature inclusion or importance and formal fairness criteria require utmost caution \citep{Cesaro.2020,Castelnovo.2022}.
The idea that only ignoring sensitive features can be sufficient to achieve formal fairness measures, often referred to as ``fairness through unawareness''~\citep{Dwork.2012}, has been shown to be mathematically flawed~\citep{kleinberg2018algorithmic}.
Some prior work interprets low feature importance of sensitive features as a guarantee for demographic parity (e.g., \citep{Jain.2020}).
However, a low feature importance of sensitive features is neither a necessary nor a sufficient criterion to satisfy formal group fairness metrics due to redundant encoding \citep{Miroshnikov.2022, Dimanov.2020} or differing base rates \citep{Dwork.2012, Lipton.2018b}.
Redundant encoding (i.e., correlations between sensitive features and other ``task-relevant'' features) is problematic because many popular post-hoc XAI methods, such as LIME and SHAP, do not account for correlations.
Further, when demographic groups have differing base rates (e.g., higher average scores in credit risk scoring), achieving demographic parity would require the decision to account for these group inequalities---which would justify the purposeful use of sensitive features~\citep{Dwork.2012}.

\paragraph{\textbf{Critique: Narrow and unacknowledged normative grounding}} \label{subsub:individual}

We further question the normative grounding of using feature inclusion or feature importance to allegedly report on fairness. 
Prior work has interpreted low feature importance of sensitive features as a form of ``fairness through unawareness'' (e.g., \citep{Ignatiev.2020}).
This approach not only disregards the possibility of redundant encoding but also makes implicit normative assumptions about the world.
As noted by \citet{Binns.2020}, this conception of fairness is based on the assumption that any group differences in features stem from ``personal choices,'' as opposed to structural injustice.
These assumptions also become paramount when contrastive explanations are employed to test whether similar individuals are receiving differing outcomes based on their sensitive features (e.g., \citep{Black.2020, Fan.2022}), because such approaches disregard any relationships between sensitive and other features.

Causality-based and correlation-based XAI methods (e.g., \citep{Datta.2016, Grabowicz.2022}) that aim to resolve the problem of redundant encoding and reveal statistical relationships between sensitive and other features also have implicit normative grounding.
In particular, they adopt an ideal mode of theorizing~\citep{fazelpour2020algorithmic}, in which they deem something as unjust if it would be unjust in an ideal world.~\citet{fazelpour2020algorithmic} have advocated for an inclusion of non-ideal perspectives in our conceptualization of algorithmic fairness, which requires engaging with the ways in which the current world is unfair and acknowledging that what would be fair in an ideal world may not be fair in the face of current injustices. 
Thus, using the information captured in sensitive features is not always morally wrong~\citep{Nyarko.2021}.
For example, consider the use of standardized tests to predict college success: the financial ability to retake the exam and access training means that the relationship between test scores and target outcomes may vary across demographic groups.
Thus, including sensitive features may, in fact, be \textit{desirable} from a fairness perspective~\citep{kleinberg2018algorithmic}.
These caveats highlight the fact that a narrow and unacknowledged normative grounding for fairness is pervasive across the literature on XAI.
It underlines the need not to misuse XAI as a ``fairness proof'' but to interpret the fairness reports based on contextual factors and normative deliberations.

\paragraph{\textbf{Critique: Narrow and unacknowledged views on formal procedural fairness}}
\citet{Balkir.2022} observe that XAI and formal fairness criteria take fundamentally different perspectives and argue that XAI tends to focus on the procedural dimension of fairness---that is, one that relates to the fairness of decision-making processes rather than outcomes~\citep{grgic2018beyond}.
Whereas \textit{formal distributive fairness} metrics (such as demographic parity and equality of opportunity) are well established in the literature and grounded on moral and political philosophy~\citep{Barocas.2019, ArifKhan.2022}, \citet{Balkir.2022} note that there is less of a shared conception of \textit{formal procedural fairness}.
Prior work has often taken the perspective that the use or disuse of sensitive features determines whether an AI system is procedurally fair or not (e.g., \citep{grgic2018beyond,wang2024procedural}).
However, this is only a narrow perspective on formal procedural fairness.
\citet{Morse.2021} propose the domain of organizational justice \citep{Colquitt.2001, Leventhal.1980} as a source for holistically defining formal procedural fairness.
Herein, \citet{Leventhal.1980} defines six components of procedural fairness: consistency, accuracy, ethicality, representativeness, bias suppression, and correctability.
For some of these components (e.g., bias suppression) XAI may in some instances be helpful; others (e.g., ethicality and correctability) would demand measures beyond formal fairness reports, such as value transparency \citep{Loi.2021} or appeal processes \citep{Lyons.2021}.
We discuss this further in \Cref{sub:judge}.

\paragraph{\textbf{Critique: Technical limitations of XAI for formal fairness reports}}
In a plea for intrinsically interpretable models, \citet{Rudin.2019} argues how model-agnostic explanations of black-box models are fundamentally unfaithful to the original model and cannot explain decision processes sufficiently.
This critique has been echoed in multiple studies demonstrating that such approaches are prone to adversarial attacks on fairness reports, which produce innocuous explanations for (formally) unfair models.
In fact, a major limitation of feature importance methods is their susceptibility to ``fairwashing'' through, for example, rationalized surrogate models \citep{Aivodji.2019}, reliance on input perturbations \citep{Slack.2020b}, or exploitation of redundant encoding \citep{Dimanov.2020}.
Some approaches claim to provide a more ``accurate'' \citep{Ghosh.2022} and ``robust'' \citep{Begley.2020} picture of the usage of sensitive features, potentially tackling issues of ``fairwashing.''
However, future work must ascertain the reliability of these analytic tools and be cautious with the interpretation of the analyses.

\citet{Rudin.2019} further questions the procedural character of many post-hoc XAI methods and argues that they do not reveal direct insights into the true underlying mechanisms of a black-box model.
Indeed, post-hoc methods like LIME, SHAP, and contrastive explanations provide little information about the decision-making process itself.
Thus, while the goal of these approaches is to report on procedural fairness, a true procedural perspective on fairness may only be provided by model-specific explanations \citep{Carvalho.2019}, such as intrinsically interpretable rule lists (e.g., \citep{Aivodji.2021b}).

\paragraph{\textbf{Critique: Power asymmetries in XAI-enabled fairness reports}}
The stakeholders (e.g., developers) in charge of producing fairness reports take a crucial role.
By making important design choices on the transparency of an AI system, they shape the way a system is perceived by other, mostly less powerful, stakeholders in downstream steps.
Many stakeholders without further access and knowledge must rely on the selective information provided by these XAI techniques.
This power dynamic is critical, especially since explanations can be manipulative~\citep{Aivodji.2019, Dimanov.2020,Lakkaraju.2020,Anders.2020}.
For XAI to become a valuable tool for fairness desiderata, it is, therefore, important to be explicit about the targeted stakeholders, their potential needs and objectives \citep{Langer.2021b}, as well as the normative deliberations that went into the development of relevant XAI techniques \citep{Loi.2021}.

\subsection{Claim 3: \textit{``XAI enables humans to analyze sources of (formal) unfairness.''}} \label{sub:analyze}
Beyond descriptive fairness reports, XAI methods are often claimed to uncover patterns of formal unfairness and to pin down contributing factors.
     \begin{quote}
     \small{
    ``The investigations demonstrate that fair decision making requires extensive contextual understanding, and AI explanations help identify potential variables that are driving the unfair outcomes'' \citep[p. 1]{Zhou.2020b}.
   }
    \end{quote}
This extends the epistemic facet of XAI to provide deeper level insights of how a specific notion of (un)fairness emerges.
Such claims concern \textit{instance-centric} or \textit{feature-centric} approaches:
instance-centric approaches focus on individual instances in the data that drive unfairness.
For example, some prior works claim to identify ``discriminatory samples''~\citep{Fan.2022} or ``discriminatory input''~\citep{Aggarwal.2019} in the training data, which refer to individual instances for which a definition of individual fairness is violated.
Feature-centric approaches analyze how features relate to formal fairness.
Some extend existing feature importance methods with the goal of quantifying the contribution of features to formal unfairness (e.g., \citep{Begley.2020, Miroshnikov.2022}).
Others causally decompose the influence of sensitive features on outcomes into direct, indirect, or induced discrimination (e.g., \citep{Zhang.2018, Grabowicz.2022}).

\paragraph{\textbf{Critique: Implicit assumptions when analyzing sources of formal unfairness}} \label{subsub:exploration}
Analyzing \emph{sources} of unfairness necessitates a definition of fairness, and as such this body of work suffers from many of the same limitations as approaches aimed to report on (formal) fairness. Unacknowledged normative grounding and misconceptions of feature importance and inclusion, discussed in \Cref{sub:report}, are also central weaknesses of existing methodologies that aim to identify sources of unfairness. To avoid repetition, we refer the reader to the previous section, and briefly highlight how these limitations apply here. 
Regarding normative grounding, instance-centric approaches heavily rely on the definition of individual discrimination and the specified similarity metric.
For example, many of these works assume that discrimination only exists when two individuals that only differ in their sensitive features receive different outcomes (e.g.~\citep{Aggarwal.2019}).
Meanwhile, applications of knowledge or causal graphs for interpretations of fairness come at the cost of making strong assumptions~\citep{chiappa2019path}, which may not always hold in practice.
Moreover, \citet{hu2020s} argue that the validity of constructing counterfactuals based on sensitive features like sex may be questionable altogether, because they make flawed ontological assumptions.
In particular, features that such diagrams understand as ``effects'' of sensitive features are in fact essential to the social meaning of said attributes.

\paragraph{\textbf{Critique: Underappreciated forms of formal fairness analysis}}
Only a few articles shed light on more subtle facets of formal fairness that emerge with AI-informed decision-making.
For example, \citet{Balagopalan.2022} and \citet{Dai.2022} argue that disparities in the quality (``fidelity'') of explanations introduce a novel kind of formal unfairness.
This adds another layer to the relationship between XAI and fairness in that explanations are not only supposed to indicate unfairness but may themselves exert a form of unfairness.
Moreover, \citet{Gupta.2019} and \citet{Karimi.2022} examine the fairness of recourse; that is, explanations that provide guidance to affected parties on how to turn a negative into a positive prediction.
Recourse can be formalized by the distance to the decision boundary, which can also introduce disparities between demographic groups that conventional fairness metrics fail to reflect.
By accounting for fairness of recourse, XAI might contribute to a more holistic view of formal fairness that not only considers discrete points in time but also addresses future actions of affected parties.
However, \citet{Slack.2021} warn that existing techniques to assess recourse typically rely on counterfactual explanations, which can be manipulated, and show how XAI methods are prone to concealing group disparities in the cost of achieving recourse (i.e., how much effort has to be invested in order to change an individual's outcome).

\subsection{Claim 4: \textit{``XAI enables humans to mitigate (formal) unfairness.''}} \label{sub:mitigate}
Several works observing formal unfairness directly employ countermeasures to mitigate it (e.g., \citep{Meng.2022}) or propose mitigation as a next step for future work (e.g., \citep{Miroshnikov.2022}).
The sequence of $(i)$ detecting, $(ii)$ analyzing sources, and $(iii)$ mitigating unfairness aligns with the distinction between epistemic and substantial facets of fairness desiderata~\citep{Langer.2021b}; where $(i)$ and $(ii)$ correspond to the epistemic facet, and $(iii)$ corresponds to the substantial facet.
In some cases, the facets can coincide when XAI methods like feature importance are directly integrated into training and bias mitigation algorithms.
    \begin{quote}
    \small{
    \textit{``To inhibit discrimination in algorithmic systems, we propose to nullify the influence of the protected attribute on the output of the system, while preserving the influence of remaining features''}~\citep[p. 1]{Grabowicz.2022}.
    }
    \end{quote}
There are several studies that employ XAI methods with the goal of mitigating formal unfairness at the pre-processing, in-processing, or post-processing stage of AI systems (e.g., \citep{Hickey.2021,Pradhan.2022,Aivodji.2021b}).
One common approach of using XAI for unfairness mitigation is to implement ``interpretable'' fairness constraints during model training, which has been done for rule lists (e.g., \citep{Aivodji.2021b}), random forests (e.g., \citep{Aghaei.2019}), and deep neural networks (e.g., \citep{Wagner.2021}).
Other methods include retraining algorithms that incorporate a fairness regularization term, which prior works compute with SHAP (e.g., \citep{Hickey.2021}) or constructed counterfactuals (e.g., \citep{Dash.2022b}).
Aiming to resolve formal unfairness arising through concept drift, others have established a monitoring system that is claimed to automatically detect formal unfairness, attribute it to a responsible feature, and mitigate it~\citep{Ghosh.2022}.
Lastly, several articles \citep{Alves.2020, Alves.2021, Alves.2021b, Bhargava.2020} propose feature dropout algorithms as a mitigation technique once an XAI method (e.g., LIME) detects reliance on sensitive features.

\paragraph{\textbf{Critique: Implicit assumptions when mitigating formal unfairness}} \label{par:deont}
Many XAI-based unfairness mitigation strategies suffer from the same shortcomings discussed in \Cref{subsub:individual}, because they rely on the same normative groundings and assumptions about features. In particular, we note that many of these methods rely on the use of XAI to identify sources of fairness---and then use this information in the mitigation step. 
For example, when relying on LIME or SHAP to reduce the feature importance of sensitive features (e.g., \citep{Alves.2020, Alves.2021b, Bhargava.2020}, redundant encoding is a salient concern\citep{Miroshnikov.2022}.
Further, employing XAI to improve measures of individual fairness should be thoroughly grounded on normative assumptions about the source of existing group disparities \citep{Binns.2020}.

\paragraph{\textbf{Critique: Risk of exacerbating unfairness}} 
The lack of explicit normative grounding also risks \emph{exacerbating} bias inadvertently instead of mitigating it. Recall the discussion in \Cref{sub:report}, where we note that using the information captured in sensitive features is not always undesirable from a fairness perspective~\citep{kleinberg2018algorithmic,Nyarko.2021}.
Group-specific differences in the predictive relationship between covariates and outcomes are known as \textit{differential subgroup validity}~\citep{hunter1979differential}.
This is common across domains and can emerge for different reasons.
For instance, previous patterns of discrimination may lead to differential predictive relationships between standardized test scores and college success~\citep{kleinberg2018algorithmic}.
In other cases, the nature of the problem may inherently lead to group-specific relationships.
For instance, the language in hate speech is different when the target group varies, and accounting for this heterogeneity can improve overall performance and fairness of hate speech detection algorithms~\citep{gupta2023same}.
In such cases, applying mitigation strategies that erode group-specific patterns could exacerbate bias stemming from differential subgroup validity by, for example, favoring the patterns that are predictive for the majority group.

\subsection{Claim 5: \textit{``XAI informs human judgment of fairness.''}} \label{sub:judge}
Whereas \Cref{sub:report} summarizes XAI methods to provide descriptive information on formal fairness, this section discusses how humans interpret this information to make (non-formal) fairness judgments.
Intuitively, if a model can justify its reasoning, a human should be able to judge whether it complies with normative standards or moral intuition.
    \begin{quote}
    \small{
    \textit{``Using XAI systems provides the required information to justify results, particularly when unexpected decisions are made. It also ensures that there is an auditable and provable way to defend algorithmic decisions as being fair and ethical, which leads to building trust''}~\citep[p. 52142]{Adadi.2018}.
    }
    \end{quote}
Stakeholders may use information generated by XAI in multiple ways.
\textit{Deployers} are often interested in using XAI to justify the decisions of their models in order to comply with legal frameworks and to foster trust and acceptance \citep{Cornacchia.2021}.
\textit{Regulators} are expected to establish and audit regulatory requirements on transparency and fairness to steer algorithmic decisions in a socially acceptable direction (e.g., \citep{Ras.2018}).
Similarly, \citet{Leslie.2019} demands deployers to prove to an external auditor that their system is ``ethically permissible, non-discriminatory/fair, and worthy of public trust/safety-securing.''
Finally, XAI aims to serve \textit{affected parties} in multiple ways.
Like other stakeholders, affected parties should be able to make well-founded judgments about model fairness, and XAI has been claimed to help in this regard (e.g., \citep{Binns.2018b}).
However, addressing the limited access to information and lack of AI literacy, it is frequently demanded that affected parties should receive a dedicated set of information to engage in an informed discourse (e.g., \citep{Russell.2017}).
These demands are reinforced by \citet{Wachter.2017}, calling for explanations that enable the understanding of decisions, support contestability, and provide guidance on recourse.

\paragraph{\textbf{Critique: Disputed value of XAI for auditors}}\label{subsub:audit}
At its most general, some works argue that XAI is not required for fairness audits (e.g., \citep{Springer.2019, Warner.2021}).
When the central concern of regulators is distributive fairness, relying on human judgement to assess this may not only be unnecessary but may also be misleading~\citep{Schoeffer.2022b}.
Moreover, several caveats on XAI for fairness reporting discussed in \Cref{sub:report} have downstream consequences for auditing; for example, the capacity of XAI methods to intentionally or unintentionally conceal the feature importance of sensitive features~\citep{Lakkaraju.2020}.
``Fairwashing'' loopholes \citep{Dimanov.2020, Aivodji.2019} pose threats to validity and robustness of expert audits.
Recent work has also alluded to similar problems in the light of model multiplicity~\citep{black2022model,marx2020predictive}; that is, situations where different predictive models are admissible.
In such cases, different models may entail different explanations, potentially allowing system designers to strategically obscure fairness concerns~\citep{black2022model}.
Generally, an alignment of normative grounding is crucial to the use of XAI for auditing.
The lack of specificity of normative grounding discussed in \Cref{sub:report} risks the misuse of XAI methods in audits where the normative grounding of the regulation is misaligned with the implicit grounding of the methodology deployed. 

\paragraph{\textbf{Critique: Disputed value of XAI for affected parties}}
Addressing informational needs of affected parties, XAI is often seen as a valuable tool in the contexts of the ``right to explanation'' (e.g., \citep{Hamon.2022,Goodman.2017}), contestability (e.g., \citep{Walmsley.2021}), and recourse (e.g., \citep{Gupta.2019}).
These desiderata are closely related to the concept of \textit{informational fairness}, which has been defined as providing ``adequate information on and explanation of the decision-making process and its outcomes''~\citep{Schoeffer.2022c}, but also to the concept of ``informed self-advocacy''~\citep{Vredenburgh.2022}.
Prior work has observed that certain types of explanations can support humans in acknowledging formal unfairness but also stresses that the context of deployment is a crucial moderator~\citep{Binns.2018b,Dodge.2019}.
Concerningly, to date, there is deficient guidance on how to design XAI for informational fairness and informed self-advocacy.
\citet{Asher.2022,Wachter.2017}, and \citet{Watson.2021} provide conceptual starting points for formal requirements of explanations for affected parties, which, to our knowledge, have not been adequately tested in practice.
Moreover, \citet{Schoeffer.2022c} find that in some cases, people have no concerns with opaque decisions and raise questions about the actionability of different explanations regarding recourse.
\citet{Schlicker.2021b} find no effects of XAI on informational justice measures, either.
It remains an open challenge to understand what kind of situational information XAI can provide to affected parties in order for them to \textit{be} and \textit{feel} treated ``fairly'' beyond the (currently) shallow concept of the ``right to explanation''~\citep{Wachter.2017}.

\paragraph{\textbf{Critique: XAI can mislead human judgment}}
Adding to the above critiques, prior research stresses that information can be communicated in \textit{unfair} ways.
\citet{Aivodji.2019} argue that the lack of specificity in XAI requirements creates incentives to provide deceptive explanations.
\citet{LeMerrer.2020} formally show that remote explainability (i.e., indirect access to a single local explanation) makes it impossible for individuals to detect untruthful manipulations.
\citet{JohnMathews.2022} proposes the concept of denunciatory power, which describes an explanation's capacity to reveal an ``unfair'' incident.
Their experimental study shows how system providers are incentivized to select the explanation with the lowest denunciatory power to minimize negative feedback.
On the other hand, misleading explanations can even occur despite benevolent intentions~\citep{Ehsan.2021b}.
Providing a game-theoretic framework, \citet{Watson.2021} conceptualize accuracy, simplicity, and relevance as the key properties of fair explanations.
That is, the provided information should be trustworthy, understandable, and helpful to the explainee.
They formally show that even accurate explanations can be misleading if they are incomprehensible or the information content is worthless.

\subsection{Claim 6: \textit{``XAI improves human perceptions of fairness.''}} \label{sub:perceived}

Beyond formal fairness, XAI is often touted to promote positive opinions and feelings about fairness of AI systems, which is closely connected with trust and acceptance (e.g., \citep{ShulnerTal.2022, Papenmeier.2019}).
    \begin{quote}
    \small{
    \textit{``The aim of local explanations is to strengthen the confidence and trust of users that the system is not (or will not be) conflicting with their values, i.e. that it does not violate fairness or neutrality''}~\citep[p. 5]{Ras.2018}.
    }
    \end{quote}
In a recent survey, \citet{Starke.2022b} find ``tentative evidence that explanations can increase perceived fairness'' and note that fairness perceptions are moderated by a range of factors, including the context of deployment, political ideology, AI literacy, and self-interest.
To disentangle the effect of XAI on perceived fairness, several studies build on the justice constructs of \citet{Colquitt.2001}, which decomposes fairness perceptions into an informational, procedural, and distributive dimension.
Accordingly, prior work has suggested that explaining models (\textit{perceived informational fairness}) enables and moderates fairness judgments about the underlying process (\textit{perceived procedural fairness}) and its outcome (\textit{perceived distributive fairness})~\citep{Binns.2018b}. 
There is some evidence that XAI is effective in increasing informational fairness and trustworthiness perceptions, even over explanations provided by human decision-makers~\citep{Schoeffer.2022c, Schoeffer.2022d}.
However, findings on perceived procedural and distributive fairness are mostly inconclusive (e.g., \citep{Binns.2018b, Schlicker.2021b}).
This might be due to the dual effect of XAI on perceived fairness described by \citet{Lee.2019b}: XAI can contribute to more understanding and transparent treatment (which relates to informational fairness); at the same time, XAI can unveil properties of the model that might conflict with people's fairness beliefs (which relates to procedural or distributive fairness).

\paragraph{\textbf{Critique: Societal concerns with maximizing perceived fairness}} \label{subsub:perceived}
Positive fairness perceptions may, in several cases, be desirable but can emerge for questionable reasons.
For example, \citet{ShulnerTal.2022b} find that the effect of explanations on perceived fairness is primarily dominated by outcome favorability.
Moreover, negative outputs tend to be regarded as unfair, regardless of the explanation \citep{ShulnerTal.2022b}.
\citet{Shin.2021b} finds that the mere act of providing explanations positively affects source credibility, which makes humans prone to form trust based on placebic~\citep{eiband2019impact} or manipulative explanations \citep{Lakkaraju.2020, Aivodji.2019}.
Similarly, it has been shown that explanations can increase participants' trust and fairness perceptions even if the scenario primes the model as unfair~\citep{Angerschmid.2022}.
From a societal perspective, this is concerning because users might inappropriately rely on unfair model output, and affected parties might not recognize that they are treated unfairly.
Therefore, a key desideratum of XAI in many cases may not be to foster \textit{positive} fairness perceptions but \textit{appropriate} (i.e., calibrated) fairness perceptions~\citep{Schoeffer.2021} instead.

\subsection{Claim 7: \textit{``XAI enables humans to implement subjective notions of fairness.''}} \label{sub:subjective}
It has been claimed that stakeholders can adjust a model towards non-formalized notions of fairness based on factors such as morale, domain-specific expertise, or other contextual factors.
Throughout a series of co-design workshops, \citet{Stumpf.2021} especially highlight users and affected parties as key stakeholders to mitigate unfairness by incorporating feedback into the model.
From the perspective of users, \citet{Schoeffer.2022b} suggest that XAI should enable humans to calibrate their reliance behavior accordingly, but also identify a lack of empirical evidence for this.
An exemplary claim in support of this idea is \citet{Chakraborty.2020} proposing an XAI method that visualizes the nearest neighbors of an unfairly classified data point.
    \begin{quote}
    \small{
    \textit{``We generate this tabular explanation for all test data points which are unfairly treated. A domain expert can easily evaluate our explanations and take decision whether to change the prediction or not''}~\citep[p. 1231]{Chakraborty.2020}.}
    \end{quote}
Moreover, XAI has been claimed to enable domain experts to make better trade-off decisions, for example, between fairness and accuracy (e.g., \citep{Aivodji.2021, Ahn.2020}).
Some works have also proposed to have users directly incorporate domain-specific interpretable constraints into the model (e.g., \citep{Zhang.2020}).
Others support the idea of actively integrating XAI-based feedback on fairness from affected parties into the design process of a model (e.g., \citep{Floridi.2020, Stumpf.2021}).
Finally, prior work has employed XAI to explore how affected parties can identify unfair use of features that developers might overlook or directly implement feedback themselves (e.g., \citep{Nakao.2022, vanBerkel.2019}).

\paragraph{\textbf{Critique: Threat of ineffective ``humans-in-the-loophole''}}
Existing laws and regulations (e.g., the GDPR), assign an essential role to a human-in-the-loop as a safeguard for fairness and accountability \citep{Hamon.2022}.
In fact, prior work has argued that human points of contact are valuable for a sense of interpersonal fairness \citep{Colquitt.2001}.
Also, human discretion may be required to make normative trade-offs \citep{Selbst.2018} and to overrule intolerable outputs \citep{Walmsley.2021}.
However, humans engaging in AI-informed decision-making should be provided with adequate tools to foster effective and responsible reliance behavior \citep{Schoeffer.2022e, Baum.2022}.
Otherwise, real-world applications might be at risk of installing ineffective ``humans-in-the-loophole'' \citep{Binns.2022} that legitimate whatever the underlying logic of the model dictates.
There is a need for effective XAI tools to capitalize on the complementary capabilities of humans and AI systems for fairness desiderata---for example, in cases where AI systems are superior in analyzing statistical patterns, and human discretion balances fairness desiderata based on the societal context.
However, empirical evidence suggests that the practical benefits of existing XAI methods for the human-in-the-loop are limited~\citep{Schoeffer.2022b}.
Instead, prior work has shown that XAI may often lead to situations where humans under- or over-rely on AI advice~\citep{schemmer2023appropriate}.

\paragraph{\textbf{Critique: XAI is not providing relevant cues to human stakeholders}}
XAI techniques need to be designed to offer meaningful cues to human stakeholders, empowering them to effectively exercise their discretionary power in pursuit of specific objectives.
With respect to formal distributive unfairness, recent work highlights concerns that widely used XAI methods such as LIME and SHAP may not provide such cues~\citep{Schoeffer.2022b}.
Instead, these techniques merely indicate whether an AI system utilizes sensitive features, which is not a reliable signal for fairness because of redundant encoding~\citep{Miroshnikov.2022, Dimanov.2020} or differing base rates~\citep{Dwork.2012, Lipton.2018b}, as discussed in \Cref{sub:report}.
Emerging XAI techniques should empower human stakeholders to base their discretion on information that is both relevant and reliable for enhancing fairness objectives.

\section{Three Patterns of Critique}
According to our survey, there is a prevailing optimism in recent literature regarding XAI as a catalyst for promoting fairness in AI-informed decision-making.
However, we argue that several of these optimistic expectations are misplaced.
Specifically, many claims on alleged fairness benefits of XAI exhibit three distinct types of shortcomings.

First, despite being highly optimistic, we find that many claims on the relationship between XAI and fairness are \textbf{vague and simplistic}.
In \Cref{subsub:futility}, we have seen that prior work has claimed XAI to be a necessary or even sufficient condition for fairness.
However, such blanket claims are prone to promoting a misguided reliance on XAI for fairness.
While prior work has shown that XAI might in some cases contribute to a better understanding of AI systems~\citep{lim2009and}, it is questionable why explainablity would be a ``precondition for ensuring their safety and fairness'' \citep{Leslie.2019} if we, for example, are only interested in fair outcomes.
Such claims may sometimes be based on an implicit assumption that XAI can enable humans to audit or rectify AI systems.
However, this gives rise to another limitation concerning a mismatch between the anticipated benefits and the actual capabilities of XAI techniques---which we will delve into shortly.
Further, the idea of ``ensuring'' \citep{Leslie.2019} or ``guaranteeing''~\citep{BarredoArrieta.2020} fairness often fails to consider the multidimensional and conflicting nature of fairness~\citep{Friedler.2019, Mulligan.2019,DeArteaga.2022}.
A one-size-fits-all fairness notion simply does not exist~\citep{Kleinberg.2017,Chouldechova.2017}.
We hope that our work can cultivate a more nuanced language for future research on the potential capabilities of XAI for fairness.

Second, many fairness desiderata pursued with XAI methods are \textbf{lacking normative grounding}.
For example, several articles treat ``reliance on sensitive features''~\citep{Alves.2020} as a form of unfairness without offering a normative rationale for why such reliance might be problematic.
\citet{Binns.2020} argues that this notion relates to a very confined form of unfairness that assumes a worldview where group disparities are solely due to personal choices.
While this view might be valid in (hypothetical) societies where no structural disadvantages occur, it is crucial to critically reflect upon this fundamental assumption~\citep{fazelpour2020algorithmic}.
In fact, prior work has shown that actively \textit{considering} sensitive features, such as gender or race, may significantly improve performance for historically marginalized groups like Black people and women and reduce algorithmic bias~\citep{kleinberg2018algorithmic,Lipton.2018b,mayson2018bias,pierson2020large,skeem2016gender}; and this is closely connected to the presence of differential subgroup validity~\citep{hunter1979differential}.
Finally, claims suggesting XAI as a means to improve fairness perceptions often fail to explain why positive fairness perceptions are a desirable goal in themselves---particularly in light of prior work showing that human perceptions are easily misled~\citep{Lakkaraju.2020,chromik2019dark,Slack.2020b}.
Instead, in many cases, a more ethically justifiable goal might be to foster \textit{appropriate} fairness perceptions, which are positive if and only if the underlying AI system is fair~\citep{Schoeffer.2021}.

Third, even in cases of specifying and motivating a valid fairness desideratum, some claims are \textbf{poorly aligned with the actual capabilities of XAI}.
For example, if the goal is to achieve formal distributive fairness, it is unclear how exactly XAI should promote this.
In fact, prior research has shown that popular feature-based explanations like LIME and SHAP are unreliable mechanisms for enabling humans to enhance formal distributive fairness through leveraging their discretionary power~\citep{Schoeffer.2022e}.
It has also been shown that many post-hoc XAI methods are not suitable for auditing AI systems due to their susceptibility to manipulations \citep{Aivodji.2019, Anders.2020, Dimanov.2020, Slack.2020b, Slack.2021}.
For example, \citet{Dimanov.2020} show that adversarial model explanation attacks can reduce the feature importance of sensitive features and still exacerbate distributive unfairness.
In some cases, XAI appears to be suited for reflecting upon the legitimacy of features (e.g., \citep{Anders.2020}), which, again, requires proper normative deliberations that go beyond what current XAI techniques alone can offer.

\section{Conclusion and Outlook}
We conducted a critical survey organizing and scrutinizing claims about the alleged fairness benefits of XAI.
Despite many optimistic positions on XAI in the recent literature, we notice that the claimed fairness desiderata are often $(i)$ vague and simplistic, $(ii)$ lacking normative grounding, or $(iii)$ poorly aligned with the actual capabilities of XAI.
To facilitate a meaningful debate and move the field forward, our work stresses the importance of embedding XAI in a sociotechnical decision-making context that considers normative motivations and societal circumstances.
Concretely, we emphasize the need to be more specific about \textbf{\textit{what}} kind of XAI method is used and \textbf{\textit{which}} fairness desideratum it refers to, \textbf{\textit{how}} exactly it enables fairness, and \textbf{\textit{who}} is the stakeholder that benefits from XAI.
We hope that future work will build on our survey to address and overcome the fairness-related limitations of XAI.

By structuring the debate and building on the differentiation of fairness dimensions from prior literature, we hope to inspire a more nuanced and precise language for future research.
A distinction between formal metrics and human perceptions helps to keep in mind that formal fairness criteria are often only meaningful when put into context and related to a moral principle.
This is especially relevant for XAI-informed fairness reports, where the use of sensitive features can indicate discrimination but can just as well be normatively justified.
The distinction between epistemic and substantial facets of XAI~\citep{Langer.2021b} clarifies the role XAI is meant to play: are XAI methods used to \textit{explore} unfairness---or do we employ them to directly \textit{alter} the model's fairness properties?
Finally, the distinction between informational, procedural, and distributive fairness~\citep{Colquitt.2001} allows a more nuanced look at specific aspects of an entire decision-making process.
For example, informational fairness can provide an interesting perspective on XAI with a focus on \textit{fair explanations} as opposed to fair decisions.
Ultimately, we encourage future work to center around our raised questions of \textbf{\textit{what}}, \textbf{\textit{which}}, \textbf{\textit{how}}, and \textbf{\textit{who}} in order to map opportunities of XAI for different human stakeholders along the full lifecycle of AI systems.

\begin{acks}
This research was supported in part by NIH grant R01NS124642, by a Google Award for Inclusion Research, and by \textit{Good Systems}, a UT Austin Grand Challenge to develop responsible AI technologies. The authors are grateful for the ongoing support from Gerhard Satzger and the KSRI team at the Karlsruhe Institute of Technology.
\end{acks}

\bibliographystyle{ACM-Reference-Format}
\bibliography{refs}


\appendix

\section{Methodologies of Surveyed Literature} \label{sub:overview}

Consistent with \citet{Langer.2021b}, the literature identified from our systematic review is highly diverse with respect to methodologies, pursued desiderata, and addressed stakeholders.
To provide an overview of the examined set of articles, we briefly summarize the methodologies used in these articles.
The key goal of this is not a perfectly distinct categorization but rather an emphasis on the types of evidence for the respective claims.
\Cref{tab:method} breaks down the methodologies used in the 175 reviewed articles and provides prominent examples to clarify the categories.
Note that the counts add up to more than 175 due to some articles using more than one method.
For example, \citet{Ahn.2020} propose a design framework, instantiate it on real-life data, and additionally conduct user studies to demonstrate its use for practitioners.

\begin{table}[h]
    \caption{Methodologies used in the 175 reviewed articles.} \label{tab:method}
    \centering
    \begin{tabular}{l c l}
    \toprule
    \textbf{Methodology} & \textbf{Count} & \textbf{Exemplary articles} \\
    \midrule
    \textbf{Conceptual} & \textbf{76}\\
    \quad\quad Framework & 35 & \citep{Floridi.2018,Kleinberg.2019,Langer.2021b}\\
    \quad\quad Argumentation & 24 & \citep{Kroll.2017,Lipton.2018,Rudin.2019}\\
    \quad\quad Literature review & 20 & \citep{BarredoArrieta.2020, DoshiVelez.2017,Lepri.2018}\\
    \textbf{ML evaluation} & \textbf{84}\\
    \quad\quad XAI/fairness method & 63 & \citep{Datta.2016, Grabowicz.2022, Zhang.2018}\\
    \quad\quad Case study & 12 & \citep{Gill.2020,Kung.2020,Miron.2021}\\
    \quad\quad Applied framework & 9 & \citep{Ahn.2020, Hardt.2021, Sharma.2020}\\
    \textbf{Human subject studies} & \textbf{29}\\
    \quad\quad Quantitative study & 23 & \citep{Binns.2018b,Dodge.2019,JohnMathews.2022}\\
    \quad\quad Qualitative study & 14 & \citep{Dodge.2019,Lee.2019b,Schoeffer.2021c}\\
    \bottomrule
    \end{tabular}
\end{table}

\textit{Conceptual} contributions comprise literature reviews and argumentation (such as position papers) building on prior work and reasoning.
Elaborate recommendations for design, evaluation, or regulation, as well as conceptual or formal models also fall into this category, labeled as frameworks.
\textit{ML evaluation} work comprises all studies that empirically evaluate a machine learning (ML) method or framework on real-world datasets.
The most prevalent type of research is the empirical evaluation of an XAI and/or fairness method.
This also includes work that scrutinizes existing XAI methods by performing adversarial attacks.
Case studies applying existing methods in a specific domain or context are also included in this category. 
Further, if a framework is empirically evaluated on data, it additionally appears in this category as applied framework. 
Finally, \textit{human subject studies} involve empirical examination of human perceptions, needs, or feedback.
While quantitative methods mostly test statistical significance of hypotheses on fairness perceptions, qualitative studies typically explore reasoning and opinions of various stakeholders.

\section{References for Archetypal Claims}\label{sub:claim_overview}
\Cref{tab:claim_overview} provides a comprehensive overview of references organized along archetypal claims, distinguishing between supportive and critical perspectives towards these claims.


\begin{table*}[t]
    \small
    \caption{Overview of references for all seven archetypal claims.}\label{tab:claim_overview}
    \begin{tabularx}{\textwidth}{>{\hsize=0.7\hsize}X >{\hsize=0.3\hsize}X >{\hsize=2.0\hsize}X} 
    \toprule
    \textbf{Archetypal claim} & \textbf{Stance} & \textbf{References} \\
    \midrule
    \multirow{2}{=}{\textit{``XAI helps achieve (a generic notion of) fairness.''}}
    & \textbf{Supportive}
    & \citep{Alufaisan.2021, Leslie.2019, NassihRym.2020, Sokol.2019, Steging.2021, Abdollahi.2018, Adadi.2018, BarredoArrieta.2020, Calegari.2020b, Ferreira.2020, Gill.2020, Vieira.2022, Calegari.2020, Cath.2018, Colaner.2022, Galinkin.2022, Sartori.2022}
    \\
    & \textbf{Critical}
    & \citep{Langer.2021b, Mittelstadt.2016}\\
    \midrule
    \multirow{2}{=}{\textit{``XAI enables humans to report on (formal) fairness.''}}
    & \textbf{Supportive} 
    & \citep{Abdollahi.2018, BarredoArrieta.2020, Berscheid.2019, deGreeff.2021, DoshiVelez.2017, Du.2021, Gilpin.2018, Hacker.2022, Hind.2019, Hohman.2019, Jain.2020, Lipton.2018, Meng.2022, Michael.2019, Papenmeier.2019, Quadrianto.2019, Ras.2018, Rosenfeld.2019, Rudin.2019, Borrellas.2021, Franke.2022, Gill.2020, Langer.2021b, Mittelstadt.2016, Sokol.2019, Vieira.2022, Aggarwal.2019, Alikhademi.2021, Alves.2020, Alves.2021b, Anders.2022, Begley.2020, Black.2020, Castelnovo.2021, Cesaro.2020, Dash.2022b, Datta.2016, Fan.2022, Franco.2022, Galhotra.2021, Grabowicz.2022, Hall.2017, Hardt.2021, Ignatiev.2020, Manerba.2021, Miron.2021, Miroshnikov.2022, Ortega.2021b, Panigutti.2021, Parafita.2021, Petrovic.2022, Qureshi.2020, Raff.2018, Sharma.2020, Stevens.2020, Szepannek.2021, Tolan.2019, Tramer.2017, Vannur.2021, Wang.2022, Zhang.2021}
    \\
    & \textbf{Critical}
    & \citep{Balkir.2022, Grabowicz.2022, Meng.2022, Kroll.2017, Seymour.2018, Aivodji.2021, Alikhademi.2021, Begley.2020, Dimanov.2020, Hall.2017}\\
    \midrule
    \multirow{2}{=}{\textit{``XAI enables humans to analyze sources of (formal) fairness.''}}
    & \textbf{Supportive} 
    & \citep{Abdollahi.2018, Aggarwal.2019, Alikhademi.2021, BarredoArrieta.2020, Du.2021, Leslie.2019, Miron.2021, Panigutti.2021, Pradhan.2022, Siering.2022, Tolan.2019, Ahn.2020, Dai.2022, Gill.2020, Miroshnikov.2022, Zhou.2020b, Anders.2020, Balagopalan.2022, Begley.2020, Black.2020, Chakraborty.2020, Chung.2020, Datta.2016, Fan.2022, Galhotra.2021, Ge.2022, Ghosh.2022, Ghosh.2022b, GonzalezZelaya.2019, Grabowicz.2022, Manerba.2021, Mutlu.2022, Ortega.2021b, Pan.2021, Quadrianto.2019, Qureshi.2020, Raff.2018, Robertson.2022, Slack.2020b, Tramer.2017, Vannur.2021, Zhang.2018, Zhang.2021, Zheng.2022}
    \\
    & \textbf{Critical}
    & \citep{Warner.2021, Alikhademi.2021}\\
    \midrule
    \multirow{2}{=}{\textit{``XAI enables humans to mitigate (formal) unfairness.''}}
    & \textbf{Supportive} 
    & \citep{Bhatt.2021, deGreeff.2021, Franke.2022, Hall.2019, Hind.2019, Jain.2020, Langer.2021b, Linardatos.2020, Meng.2022, Miroshnikov.2022, Siering.2022, Tolan.2019, Wagner.2021, Zhang.2018, Aghaei.2019, Aivodji.2021, Alves.2020, Alves.2021, Alves.2021b, Bhargava.2020, Dash.2022b, Fan.2022, Ge.2022, Ghosh.2022b, Grabowicz.2022, Gupta.2019, Hall.2017, Hickey.2021, Kehrenberg.2020, Lawless.2021, Mutlu.2022, Pan.2021, Parafita.2021, Pradhan.2022, Russell.2017, Wang.2020d, Yang.2020, Zhang.2020, Zheng.2022}
    \\
    & \textbf{Critical} 
    & \citep{Franke.2022, Waller.2022}\\
    \midrule
    \multirow{2}{=}{\textit{``XAI informs human judgement of fairness.''}}
    & \textbf{Supportive} 
    & \citep{Adadi.2018, Berscheid.2019, Binns.2018b, Cath.2018, Chan.2022, Cornacchia.2021, deFineLicht.2020, Dodge.2019, DoshiVelez.2017, Gill.2020, Hacker.2022, Hamon.2022, Hind.2019, Kizilcec.2016, LeMerrer.2020, Loi.2021, Maclure.2021, Papenmeier.2022, Raff.2018, Russell.2017, Shin.2021, Sokol.2020, vanBerkel.2021, Wachter.2017, Watson.2021, Zucker.2020, Asher.2022, Floridi.2018, Franke.2022, Gupta.2019, Hall.2019, Hickey.2021, JohnMathews.2022, Karimi.2022, Lee.2019b, Leslie.2019, Loi.2021, P.2021, Schoeffer.2022c, Seymour.2018, ShulnerTal.2022b, Springer.2019, Warner.2021, Galhotra.2021, Sharma.2020, Angerschmid.2022, Anik.2021, Grimmelikhuijsen.2022, Nakao.2022}
    \\
    & \textbf{Critical} 
    & \citep{Aivodji.2019, Begley.2020, Chakraborty.2020, Chan.2022, Dai.2022b, deFineLicht.2020, Galinkin.2022, Gill.2020, Gilpin.2018, Grimmelikhuijsen.2022, Gryz.2020, JohnMathews.2022, Herman.2017, Kasirzadeh.2021, Kroll.2017, LeMerrer.2020, Lepri.2018, Loi.2021, Maclure.2021, Mittelstadt.2016, Rudin.2019, Schoeffer.2021, Schoeffer.2022b, Selbst.2018, Seymour.2018, Shin.2021b, Slack.2020b, ShulnerTal.2022, Springer.2019, Wachter.2017, Walmsley.2021, Warner.2021, Watson.2021, Anders.2020, Angerschmid.2022, Balagopalan.2022, Binns.2018b, Dimanov.2020, Slack.2021}\\
    \midrule
    \multirow{2}{=}{\textit{``XAI improves human perceptions of fairness.''}}
    & \textbf{Supportive} 
    & \citep{Papenmeier.2019, Ras.2018, Schoeffer.2021, Shin.2021b, ShulnerTal.2022, ShulnerTal.2022b, Angerschmid.2022, Anik.2021, Binns.2018b, Dodge.2019, JohnMathews.2022, Lee.2019b, Park.2021, Park.2022, Schoeffer.2021c, Schoeffer.2022c, Schoeffer.2022d, Shin.2021, Shin.2022b, vanBerkel.2021, Wang.2020, Starke.2022}
    \\
    & \textbf{Critical} 
    & \citep{Binns.2018b, Lee.2019b, Rader.2018, Schlicker.2021b, Schoeffer.2022c, Shin.2022b, Starke.2022}\\
    \midrule
    \multirow{2}{=}{\textit{``XAI enables humans to implement subjective notions of fairness.''}}
    & \textbf{Supportive} 
    & \citep{Ahn.2020, Aghaei.2019, Chan.2022, deGreeff.2021, Dodge.2019, Floridi.2020, Maclure.2021, Stumpf.2021, Wagner.2021, Zhang.2021, Zucker.2020, Aivodji.2021, Chakraborty.2020, Kung.2020, Panigutti.2021, Slack.2020, Yang.2020, Zhang.2020, Nakao.2022, Park.2022, vanBerkel.2019}
    \\
    & \textbf{Critical} 
    & \citep{Nakao.2022, Schoeffer.2022b}\\
    \bottomrule
    \end{tabularx}
\end{table*}

\end{document}